\documentclass{svmult}

\usepackage{amsmath,amsfonts}
\usepackage{cleveref}
\usepackage{float}
\usepackage{geometry}
\usepackage[cal=boondox]{mathalfa}
\usepackage{mathtools}
\usepackage{multirow}
\usepackage{indentfirst}
\usepackage{ctable}
\usepackage{titlesec}
\usepackage{times} 
\usepackage{xcolor}

\DeclareMathAlphabet{\mathbmit}{OML}{cmm}{b}{it}
\DeclareMathAlphabet{\pazocal}{OMS}{zplm}{m}{n}

\DeclarePairedDelimiter\parentheses{\lparen}{\rparen}

\renewcommand{\vec}[1]{\mathbmit{#1}}
\let\matr\vec
\newcommand{\vecHat}[1]{\widehat{\mathbmit{#1}}}
\let\matrHat\vecHat

\DeclareMathOperator*{\argmin}{arg\,min}
\DeclareMathOperator{\sign}{sgn}
\DeclarePairedDelimiterX{\inp}[2]{\langle}{\rangle}{#1,#2}

\newcommand{\ifm}[1]{\text{if $#1$}}
\newcommand{\rank}[1]{\operatorname{rank} \parentheses*{#1}}

\newcommand{\Hquad}{\hspace{0.5em}} 

\definecolor{bluepastel}{RGB}{174,198,207}
\definecolor{militarypastel}{RGB}{119,190,119}
\definecolor{orangepastel}{RGB}{255,179,71}

\usepackage{listings}
\usepackage[ruled,linesnumbered,noend]{algorithm2e}
 
\begin{document}

\title*{Balancing Predictive Relevance of Ligand Biochemical Activities}
\author{Marek Pecha} 
\institute{
Department of Applied Mathematics, FEECS, V\v{S}B-TU Ostrava 
\and 
Czech Academy of Sciences, Institute of Geonics \email{marek.pecha@vsb.cz}
}

\maketitle

\abstract{
			In this paper, we present a technique for balancing predictive relevance models related to supervised modelling ligand biochemical activities to biological targets.  We train uncalibrated models employing conventional supervised machine learning technique, namely Support Vector Machines. Unfortunately, SVMs have a serious drawback. They are sensitive to imbalanced datasets, outliers and high multicollinearity among training samples, which could be a cause of preferencing one group over another. Thus, an additional calibration could be required for balancing a predictive relevance of models. As a technique for this balancing, we propose the Platt's scaling. The achieved results were demonstrated on single-target models trained on datasets exported from the ExCAPE database. Unlike traditional used machine techniques, we focus on decreasing uncertainty employing deterministic solvers.
}

\section{ Ligand-based Supervised Biochemical Activity Modelling}
 
	\textit{In silico} computational supervised modelling based on ligand profiles is widely used as a complementary method in the pharmaceutical industry; specifically in the early stage of drug design and development to obtain an indication of off-target interactions. The aim is to train a model that generalizes the biological activities of ligands, i.e. small biochemical molecule, ion, or protein, by reusing information from former \textit{in vitro} laboratory experiments. A common approach to predicting activities of ligands on biological targets, e.g. genes, is to exploit a panel of structure-analysis models such as quantitative structure-activity relationship (QSAR) with one model per a biological target. Then, using statistical learning, we train a classification or regression model used to predict categorical (active or inactive) or numerical values, e.g. probability of reaction, for a new unseen ligand, respectively. The suggested strategies on how to train the models are the single and multi-target schemes. In our text, we focus on a simplified single-target modelling that is a baseline for more general approaches, e.g. the multi-target ones mentioned above.

	Several machine learning methods are typically employed for constructing such models. Currently, the most popular ones are Deep Neural Network (DNNs) and Support Vector Machines (SVMs). Despite the fact that Deep Learning (DL) is getting popular in recent years, SVMs are still applicable. One of the most advantages of the SVMs approach is that they find learning functions maximizing geometric margins, unlike generally composed functions produced by DNNs. A benefit of this additional implicit (in contrast with DNNS) requirement is that reducing generalization error leading to preventing overfitting of a model and improving its robustness in terms of a bias-variance tradeoff. Respecting later, SVM seems to be appropriate as a technique for modelling biochemical activity, which exploits ligand profiles, for the reason that a multivariance among samples is typically small. This fact is arising from the nature of in vitro experiments because laboratories test biochemical activities of quite similar ligands to a particular target commonly than testing ligands belonging to entirely different groups.

	However, SVMs have a serious drawback; they are sensitive to imbalanced datasets, outliers and multicollinearies among training samples, which could be a cause of preferencing one group over another. Therefore, we propose to use Platt scalling for an additional model calibration, which is based on transforming the SVM classification model output into a posterior probability distribution by fitting logistic regression model to SVM raw prediction scores. This calibrating technique is practically used for reducing impact of overfitting to predictor mainly caused by training data. After training calibrated models, we demonstrate balanced predictive relevance of these models by converting them to label prediction using an optimal threshold.

	This text is organized as follows. In Section 2, the SVM formulations with a relaxed-bias term are presented. A calibration technique based on the Platt scaling is introduced in Section 3. Numerical experiments are presented in Section 4.
	
\section{Support Vector Machines for Relaxed-bias Classification}
\label{sec:SVM} 

SVMs belong to conventional machine learning techniques, and they are practically used for both classification and regression. Unlike the DL underlying architecture, SVMs could be considered as the single perceptron problems that find the learning functions maximizing the geometric margins. Therefore, we can explain the qualities of a learning model and the underlying solver behaviour straightforwardly. In this paper, we focus only on the linear C-SVM for classification, in which a misclassifications error term is penalized by a user-specified penalty $C$. We denote the C-SVM as SVM for simplification in further text.

	SVM was originally designed by \cite{CorVap-ML-1995} as a supervised binary classifier, i.e. a classifier that decides whether a sample falls into either Class A or Class B by means of a model determined from already categorised samples in the training phase of the classifier. Let us denote the training data as an ordered sample-label pairs such that 
	\begin{equation}
		T := \{ \left(\vec{x}_1, y_1\right), \ \left(\vec{x}_2, y_2\right), \ \dots, \ \left(\vec{x}_m,  y_m\right) \},
	\end{equation}
	where $m$ is the number of samples, $\vec{x}_i \in \mathbb{R}^n, \ n \in \mathbb{N},$ is the $i$-th sample and $y_i \in \{-1, 1\}$ denotes the label of the $i$-th sample, $i \leq m$. The linear SVM solves the problem of training the classification model in the form of a so-called maximal-margin hyperplane: 
	\begin{equation}
		\label{eq:hyperplane}
		H = \inp{\vec{w}}{\vec{x}} + \hat{b},
	\end{equation}
where $\vec{w}$ is a normal vector of hyperplane $H$ and $\hat{b} = \frac{b}{\|\vec{w}\|}$ determines the offset of the hyperplane $H$ from the origin along its normal vector $\vec{w} \in \mathbb{R}^n$. In the cases of the relaxed-bias classification, we do not consider $\hat{b}$ in a classification model, however we include it into the problem by means of augmenting the vector $\vec{w}$ and each sample $\vec{x}_i$ with an additional dimension so that $\vec{\widehat{w}} \leftarrow \begin{bmatrix}\vec{w} \\ B \end{bmatrix}$, $\vec{\widehat{x}_i} \leftarrow \begin{bmatrix}\vec{x}_i \\ \gamma \end{bmatrix}$, and $\gamma \in \mathbb{R}^+$ is a user defined variable, typically set to $1$.

	Let $p \in \{1, 2\}$ for purposes related to our application, then the problem of finding hyperplane $\widehat{H} = \inp{\vecHat{w}}{\vecHat{x}}$ can be formulated as a constrained optimization problem in the following primal formulation:
	\begin{equation}
		\label{eq:svmRelaxedBiasPrimal}
    	\argmin_{\vecHat{w}, \ \widehat{\xi}_i} \ \frac{1}{2} \inp{\vecHat{w}}{\vecHat{w}} +  \frac{C}{p} \sum_{i = 1}^n \widehat{\xi}_i^p 
      	\Hquad
      	\text{s.t.} 
      	\Hquad
        \begin{cases}
        	\ y_i \inp{\vecHat{w}}{\vecHat{x}_i} \geq 1 - \widehat{\xi}_i,
        	\\
        	 \ \widehat{\xi}_i \geq 0 \Hquad \ifm{p = 1},  \ i \in \{1,2, \dots, n\},
       \end{cases}
\end{equation}
	where  $\widehat{\xi}_i = \max\{0, 1 - y_i \inp{\vecHat{w}}{\vecHat{x}_i} \} $ is the hinge loss function releated to augmented samples $\vecHat{x}_i$. Essentially, the hinge loss function quantifies error between predicted and correct classification of sample $\vecHat{x}_i$. The variable $C \in \mathbb{R}^+$ is a penalty parameter that penalizes misclassification error. Generally, a higher value of $C$ increases the importance of minimising the hinge loss functions $\xi_i$ and, on the other hand, causes maximizing $\|\vec{w} \|$, i.e. minimizing the width of margin, leading to poor generalization capabilities of a classification model. The goal is to find a reasonable value of $C$ such that a resulting model balances the robustness and performance tradeoff.
	
		 Further, we can say in general, the minimizer associated with formulation \eqref{eq:svmRelaxedBiasPrimal}  corresponding to an optimal rotation of separating hyperplane $\widehat{H}$ in one-dimension higher feature-space $\mathbb{R}^{n + 1}$ than the original feature space has.
	
	To reduce a number of unknowns, we can dualize the primal formulation \eqref{eq:svmRelaxedBiasPrimal} using the Lagrange duality so that, for both $p = 1$ and $p = 2$, they result in:
	\begin{equation}
		\label{eq:svmDualL1RelaxedBias}
    	\argmin_{\boldsymbol{\alpha}} \ \frac{1}{2} \boldsymbol{\alpha}^T \matr{Y}^T \matr{K} \matr{Y} \boldsymbol{\alpha} - \boldsymbol{\alpha}^T \vec{e}  \Hquad \text{s.t.}  \Hquad    \vec{o} \leq  \boldsymbol{\alpha} \leq C\vec{e},
		\end{equation}
		\begin{equation}
			\label{eq:svmDualL2RelaxedBias}
     		\argmin_{\boldsymbol{\alpha}} \ \frac{1}{2} \boldsymbol{\alpha}^T \left(\matr{Y}^T \matr{K} \matr{Y}+ C^{-1} \matr{I} \right) \boldsymbol{\alpha} -  \boldsymbol{\alpha}^T \vec{e}  \Hquad \text{s.t.} \Hquad    \vec{o} \leq  \boldsymbol{\alpha},
	\end{equation}   
	respectively. $\matr{K} \in \mathbb{R}^{m \times m}$ is matrix of inner products called Gramian such that $\matr{K} := \matr{X}^T \matr{X}$, $\matr{X}= \left[\vec{x}_1, \ \vec{x}_2, \ \dots, \vec{x}_m \right]$ is data matrix of the training samples, $\vec{y} = \left[y_1, \ y_2, \  \dots, \ y_m\right]^T$ is vector of corresponding labels, $\matr{Y} = diag(\vec{y})$, and  $\vec{o} \in \mathbb{R}^m$, $\vec{e} \in \mathbb{R}^m$ denote a zero-vector and an all-ones vector, respectively. In general, the Gramian $\matr{K}$ is symmetric positive semi-definite (SPS) of a rank 
	\begin{equation}
		\rank{\matr{K}} = \min\{\min\{m,n\}, N_g\},
	\end{equation}		
	where $N_g$ maximum number of linearly independent training samples, $m$ and $n$ are a numbers of training samples and their features, respectively. 
	
		Comparing the dual SVM-QP formulations, specifically, $\mathcal{l}$1-loss \eqref{eq:svmDualL1RelaxedBias} and $\mathcal{l}$2-loss \eqref{eq:svmDualL2RelaxedBias}, we can see that they differ in the forms of the related Hessians and constraints. While the Hessian is generally an SPS matrix in a case of \eqref{eq:svmDualL1RelaxedBias}, the Hessian related to the formulation \eqref{eq:svmDualL2RelaxedBias} is regularized by means of the matrix $C^{-1} \matr{I}$. This can provide a better convergence rate for \eqref{eq:svmDualL2RelaxedBias} and an associated optimization problem could be more stable. On the other hand, $\mathcal{l}1$-loss SVM could produce a more robust model in the sense of performance score, because using a linear sum of $\xi_i$ leads to catching the outliers during a training phase of a classifier. 
	
	Further, for obtaining a solution of the original primal problem, we introduce dual to primal reconstruction formula as follows:
\begin{equation}
\label{eq:recForW}
  \vecHat{w} = \matrHat{X}\matr{Y}\boldsymbol{\alpha},
\end{equation}

	Using the reconstructed normal vector $\vecHat{w}$, we can set the decision rule:
	\begin{equation}
		\label{eq:svmDecision}
		\sign\left(\inp{\vecHat{w}}{\vecHat{x}_i}\right) =
		\begin{cases}
			\ +1 \dots \ \vecHat{x}_i \in \text{Class A,} \\
			\ -1 \dots \ \vecHat{x}_i \in \text{Class B.}	
		\end{cases}
	\end{equation}

	 In sense of equivalence of solutions, we can easily show a connection between the classification models associated with standard model \eqref{eq:hyperplane} and relaxed-bias formulations. Let us write the separating hyperplane equation in a component-wise form such that:
	\begin{equation}
  		\label{eq:svmModHyperplane}
		\widehat{H} := \inp{\vecHat{w}}{\vecHat{x}} = \underbrace{ w_1 x_1 + w_2 x_2 + \dots + w_m x_m}_{ = \inp{\vec{w}}{\vec{x}}} + \underbrace{B \gamma}_{=: b},
	\end{equation}	 
	which is equivalent to \eqref{eq:hyperplane}. However, the bias term $b$ is incorporated into the regularization term in sense of Tikhonov regularization, the resulting model could slightly differ from this attained by standard (non-relaxed) formulations. On the other side, we have not to deal with equality constraints, which are appear in the standard dual formulations, and sometimes, they are reason why solvers could diverge.
	
		\section{Model Calibration}
	
	Calibrating a classification model refers to a special type of statistical inference that transforms a uncalibrated output (raw prediction), particularly, decision function $f\!\left(\vec{x}\right)$, to a probability of class membership $P(\textit{class} \ | \ \textit{input})$. Commonly, the calibration is required when we need to adjust robustness of a classification model, balances of class preference and provides a cost-sensitive classification. In this section, we pay attention to an estimation of the probability employing a well-known calibration technique called Platt's scaling -- introduced by {\cite{Platt-Advances-1999}}. Commonly, this technique is known as Platt's calibration in the machine learning communities.
	
	An idea beyond this technique is based on fitting a parametric form of sigmoid-shaped function that maps the uncalibrated SVM output to the posterior probability $P\left(y = 1 \ | \ \vec{x}\right) \approx P_{A, B}\left(y = 1 \ | \ \vec{x}\right)$, where 
	\begin{equation}
		\label{eq:plattPosteriorProb}
		P_{A, B}\left(y = 1 \ | \ \vec{x}\right) = \frac{1}{1 + \exp\left(Af(\vec{x}) + B\right)}.
	\end{equation}
	The parameters $A$, $B$ determine the slope of the sigmoidal curve and lateral displacement, respectively, and they are practically fitted using maximum likelihood estimation (MLE). In order to the relaxed-bias classification mentioned in Section 2, we assume that the raw SVM output 
	\begin{equation}		
		f(\vec{x}) := \widehat{H}\!\left(\vecHat{x}  \leftarrow \begin{bmatrix}\vec{x} \\ \gamma \end{bmatrix} \right) = \left\langle \vecHat{w}, \vecHat{x} \right\rangle
	\end{equation}		
is proportional to the log odds of positive samples in the model \eqref{eq:plattPosteriorProb}.

	In the original paper, Platt suggested to use an additional training set, i.e. a calibration set, for training calibration curve on output of general instance-based SVM to avoid incorporating bias failures, i.e. cases when $1 - y_i f_i > 0$, on functional-margin $\gamma = | 1 |$. Let us denote such dataset as an ordered set:
\begin{equation}
T_{CA} := \{\left(f_1, y_1\right), \left(f_2, y_2\right), \dots \left(f_l, y_l\right) \},
\end{equation}
where $l$ is a number of the calibration samples, $f_j$ is estimate of $f\left(\vec{x}_j\right)$ for $j \in \{1, 2, \dots, l\}$. On the other hand, when an optimal model performance is attained in a reasonably small value of the penalty $C$, e.g., in real-world applications employing linear SVMs, see {\cite{Platt-Advances-1999}}, and data is well-behaved, bias on margin failures usually become small. Therefore, it often possible to simply fit the sigmoid on the training dataset.

	To prevent model overfitting, Platt proposed additional transformation of binary labels $y_j$ to target probabilities $t_j$ such that $t_j = \frac{N_{p} + 1}{N_{p} + 2} \ \text{iff} \ y = +1$, or $t_j = \frac{1}{N_n + 2} \ \text{iff} \ y = -1$, where $N_{p}$ and $N_{n}$ are numbers of positive and negative calibration samples, respectively. 

	The best parameter setting $\left(A^*, B^*\right)$ is determined by minimizing negative log likehood (cross-entropy error function) on calibration data so that:
\begin{equation}
\label{eq:plattOptProb}
\left(A^*, B^*\right) = \argmin_{A, B} - \sum_{j = 1}^l \left[ \  t_j \log\left(p_j\right) + \left(1 - t_j\right)\log\left(1 - p_j\right) \ \right],
\end{equation}
where $p_j = \frac{1}{1 + \exp\left(Af_j + B\right)}$. To solve \eqref{eq:plattOptProb}, an author in \cite{Platt-Advances-1999} proposed to use Levenberg–Marquardt (LM) algorithm \cite{Levenberg-QAM-1944}. Unfortunately, a technique for updating a damping factor for LM introduced by Platt causes that solver could not converge to a minimum of (12). It is discussed in \cite{Lin-ML-2007}. To avoid issue arising from a damping factor associated with the LM method, the authors suggested the Newton method with backtracking line-search. 

Though the proposed approach is favourable due to its simplicity, the trust-region methods are more robust. Since we focus on training robust predictors in this paper, we exploit the Newton method with trust region in all numerical experiments presented in Section 4.

\section{PermonSVM}
	
	The PermonSVM package is a part of the PERMON toolbox designed for usage in a massively parallel distributed environment containing hundreds or thousands computational cores. Technically, it is an extension of the core PERMON package called PermonQP, from which it inherits basic data structures, initialization routines, build system, and utilizes computational and QP transformation routines, e.g. normalization of an objective function, dualization, etc. Programmatically, core functionality of PERMON toolbox is written on the top of the PETSc framework, follows the same design and coding style, making it easy-to-use for anyone familiar with PETSc. It is usable on all main operating systems and architectures consisting of smartphones through laptops to high-end supercomputers. 

	PermonSVM supports distributed parallel (through MPI) reading from formats like SVMLight, HDF5, PETSc binary file formats,  more than 4 problem formulations of classification problem, two types of parallel cross-validation, namely \textit{k}-fold and stratified \textit{k}-fold cross-validation. The resulting QP-SVM problem with implicitly represented Hessian, in which Gram matrix $\matr{X}^T\matr{X}$ is not assembled, is proceeded by solvers provided by the PermonQP package or the PETSc framework. Unlike standard machine learning libraries, PERMON toolbox provides interface functions to change underlying QP-SVM solver, monitoring and tweaking the algorithms. In Code 1, we present an example of a usage PermonSVM API.  Our libraries are developed as an open-source project under the BSD 2-Clause Licence.
	
		\begin{lstlisting}[caption={Example of calling PermonSVM API.}, label=permonsvmapi]
MPI_comm    comm = PETSC_COMM_WORLD;
SVM         svm;
PetscViewer v;

char        f_train[PETSC_MAX_PATH_LEN] = "cnr1.h5";
char        f_test[PETSC_MAX_PATH_LEN] = "cnr1.t.h5";

TRY( SVMCreate(comm,&svm) );
TRY( SVMSetType(svm,SVMPC) ); /* Platt calibration type */
TRY( SVMSetFromOptions(svm) );

TRY( PetscViewerHDF5Open(comm,f_train,FILE_MODE_READ,&v) );
TRY( SVMLoadTrainingDataset(svm,viewer) );
TRY( PetscViewerDestroy(&v) );

TRY( PetscViewerHDF5Open(comm,f_train,FILE_MODE_READ,&v) );
TRY( SVMLoadCalibrationDataset(svm,viewer) );
TRY( PetscViewerDestroy(&v) );

TRY( PetscViewerHDF5Open(comm,f_test,FILE_MODE_READ,&v) );
TRY( SVMLoadTestDataset(svm,v) );
TRY( PetscViewerDestroy(&v) );

TRY( SVMSetHyperOpt(svm,PETSC_TRUE) );
TRY( SVMSetNfolds(svm,3) );

TRY( SVMTrain(svm) );
TRY( SVMTest(svm) );

TRY( SVMDestroy(&svm) );
	\end{lstlisting}	
	
		\section{Numerical Experiments}
	
In this section, we analyze numerical experiments related to balancing predictive relevance of the single-target relaxed-bias classification model using the Platt's Calibration technique, which we introduced in Section 2 and Section 3, respectively. We benchmark this approach on datasets associated with biochemical activities of ligands on $4$ biological targets, namely abl$1$ (Abelson murine leukemia viral oncogene homolog $1$ protein), adora$2$a (Adenosine A\textsubscript{$2$A} receptor), cnr$1$ (cannabinoid receptor type $1$), and cnr$2$ (cannabinoid receptor type $2$). These datasets were exported from the  ExCAPE database, which was developed by \cite{Sun-JOC-2017}. While it is possible to calibrate a model on the same dataset, on which the model was trained, see Section 3, it could be problematic to decide if a bias of an uncalibrated model is small enough. Therefore, we split training samples into the training and calibration datasets.  After training models, we evaluate their performance on the test dataset using precision, sensitivity, and area under the curve receiver operating characteristic (AUC) performance scores. The input datasets were divided into training, calibration and test datasets such that they consist of $640$, $200$, $160$ ligands, respectively, and a ratio of active and inactive ones is sufficiently preserved. Characteristics associated with these datasets are summarized in \Cref{tab:pocDatasets}.

\begin{table}
\centering
\caption{The characteristics of training, calibration and test dataset related to abl$1$, adora$2$a,  cnr$1$, cnr$2$ biological targets.}
\begin{tabular}{| l | r | r  | r | }
\hline
 \multirow{2}{*}{Target (dataset) \hspace{10pt}} & \multicolumn{3}{c|}{\#ligands} \\
 \cline{2-4}
 &  \multicolumn{1}{c|}{\#active} &  \multicolumn{1}{c|}{\#incative} & $\sum$ \\
\hline
abl$1$ (training) & $312 \ (48.75 \%)$ & $328 \ (51.25 \%)$ & $640$ \\ 
abl$1$ (calibration) & $92 \ (46.00 \%)$ & $108 \ (54.00 \%)$ & $200$ \\ 
abl$1$ (test) & $81 \ (50.62 \%)$ & $79 \ (49.38 \%)$ & $160$ \\
\hline
adora$2$a (training) & $343 \ (53.59 \%)$ & $297 \ (46.40 \%)$ & $640$ \\ 
adora$2$a (calibration) & $105 \ (52.50 \%)$ & $95 \ (47.50 \%) $ & $200$ \\ 
adora$2$a (test) & $95 \ (59.38 \%)$ & $65 \ (40.62 \%)$ & $160$ \\
\hline
cnr$1$ (training) & $392 \ (61. 25 \%)$ & $248 \ (38.75 \%)$ & $640$ \\
cnr$1$ (calibration) & $123 \ (61.50 \%)$ & $77 \ (38.50 \%)$ & $200$ \\
cnr$1$ (test) & $110 \ (68.75 \%)$ & $50 \ (31.25 \%)$ & $160$ \\
\hline
cnr$2$ (training) & $405 \ (63.28 \%)$ & $235 \ (36.72 \%)$ & $640$ \\
cnr$2$ (cablibration) & $127 \ (63.50 \%)$ & $73 \ (36.50 \%)$ & $200$ \\
cnr$2$ (test)  & $112 \ (70.00 \%)$ & $48 \ (36.50 \%)$ & $160$ \\
\hline
\end{tabular}
\label{tab:pocDatasets}
\end{table}

	For training uncalibrated classification models, we choose the best penalty \( C_{BE} \) from the set $\widehat{C} = \{2^p, p\in\{-7, -6, \dots,6,  7\}\}$ algorithmically employing the hyperparameter optimization (HyperOpt) by means of grid-search combined with stratified $3$-fold cross validation (CV). The value of the best penalty $C_{BE}$ is selected so that accumulated related precision and sensitivity during CV are maximized. All components of the initial guess $\vec{x}_0$ are set to $0.99 * C$, proposed in \cite{Pecha-LNEE-2019}. The relative norm of projected gradient being smaller than $1e-1$, discussed in \cite{Pecha-SVM-AIP-2018}, is used as stopping criterion for the MPRGP (Modified Proportioning and Reduced Gradient Projection) algorithm, see \cite{Dos-book-09}, in all presented experiments. The expansion step-length $\alpha$  is fixed and determined such as $\alpha = 1.95 / \| \matr{A} \|_2$, where $\| \matr{A} \|_2 = \sqrt{\lambda_{max}\left(\matr{A}^T \matr{A}\right)}$, where $\matr{A}$ denotes the Hessian matrix associated with \eqref{eq:svmDualL1RelaxedBias} and \eqref{eq:svmDualL2RelaxedBias}. 

	Using PETSc implementation of the Newton method without preconditioning with default setting, the sigmoid-shaped calibration function is computed by minimizing cross-entropy \eqref{eq:plattOptProb} on calibration data. Since the Newton method converges quickly to optimal solution $\vec{x}^*$ when vector $\vec{x}$ is close enough to $\vec{x}^*$, \cite{Platt-Advances-1999} proposed initial guesses for parameters of sigmoid such that $A_0 = 0$ and $B_0 = \log{\frac{l^+ + 1}{l^- + 1}}$, where $l^+$ and $l^-$ denote numbers of active and inactive samples associated with the calibration dataset $T_{CA}$. To avoid numerical difficulties or catastrophic cancellations that could arise from evaluation $1 - p_i$, where $p_i$ is close to $1$, we evaluate \eqref{eq:plattPosteriorProb} by using $\frac{\exp\left(-Af(\vec{x}) - B\right)}{1 + \exp\left(-Af(\vec{x}) - B\right)}$ when $Af(\vec{x}) + B \geq 0$ else we use \eqref{eq:plattPosteriorProb}. This numerical improvements were proposed by \cite{Lin-ML-2007}. Other numerical obstacles could arise from evaluating Hessian 
\begin{equation}
	\label{eq:crossEntropyHessian}
	H = 
	\begin{bmatrix}
		\sum_{i = 1}^l f_i^2p_i\left(1 - p_i\right) & \sum_{i = 1}^l f_ip_i\left(1 - p_i\right) \\
		\sum_{i = 1}^l f_ip_i\left(1 - p_i\right) & \sum_{i = 1}^l p_i\left(1 - p_i\right)
	\end{bmatrix}
\end{equation}
associated with cross-entropy function \eqref{eq:plattOptProb}. Thus, we replace the term $\left(1 - p_i\right)$ by means of $\frac{1}{1 + \exp\left(-Af(\vec{x}) - B\right)}$ when $Af(\vec{x}) + B \geq 0$, else $\frac{\exp\left(Af(\vec{x}) + B\right)}{1 + \exp\left(Af(\vec{x}) + B\right)}$, see \cite{Lin-ML-2007}. Since the Hessian 	\eqref{eq:crossEntropyHessian} is SPS in general \cite{Lin-ML-2007}, we regularize it by the matrix $\sigma\matr{I}$, where $\sigma = 1\mathrm{e}{-12}$ in our experiments.

	Instead of stochastic optimization, which commonly used in the machine learning community, our used solvers are deterministic in these experiments. They pass all training samples in one iteration. Therefore we consider terms epoch and iteration as identical in this text. In other words, we do not take into account a batch of a training dataset during the training phase of the classifier. By this, we obtain strictly settled and reproducible training pipelines, unlike employing DNNs or other traditionally used techniques. On the other hand, deterministic solvers suffer on their cost in the sense of computational resources. Thus, training predictors can take longer than in the case of stochastic optimization. At this expense, we reduce uncertainty during the training process, which could be crucial for some scientific application, e.g. ones related to the pharmaceutical industry.

	After calibrating a classification model, we convert probabilities to label prediction using optimal threshold (thr.),  i.e. $y = 1 \ \ifm{p > \text{thr.}}$ to demonstrate the balanced class predictive relevance of calibrated models. The optimal threshold is determined using grid-search so that absolute value of the difference of precision score (Pre.) and sensitivity (Sen.) on the test dataset is minimized, and F$1$ score must be greater than $0.50$, i.e. predictive ability of model must be better than random. 

	Because of all datasets are too small to utilize more than one processor core, all experiments were run on $1$ MPI process pinned to a processor core. In all presented experiments, we utilized the same node of the ANSELM supercomputer at IT4Innovations. Evaluations of performance scores are summarized in \Cref{tab:uncalibratedModelScores} and \Cref{tab:calibratedModelScores} for uncalibrated models and models after calibration, respectively.

\begin{table}[H]
	\centering
    \caption{abl$1$, adora$2$a, cnr$1$, cnr$2$ uncalibrated single-target models: evaluation of performance scores of models trained with parameter $C_{BE}$ determined by means of HyperOpt, particularly, grid-search combined with $3$-fold cross-validation. We report precision (Pre.), sensitivity (Sen.), F1 score, and AUC as qualitative metrics of a model performance for the both  $\mathcal{l}1$-loss and $\mathcal{l}2$-loss SVM.}
     \begin{tabular}{| l | c | c | c | c | c | c | c |}
        \hline
        \multirow{2}{*}{Target} & \multirow{2}{*}{\hspace{0.1pt} Loss } & \multicolumn{5}{c|}{ Uncalibrated model } \\
        \cline{3-7}
        & &  $C_{BE}$ & Pre. [\%] & Sen. [\%] & F$1$ & AUC \\
        \hline
        \multirow{2}{*}{abl$1$} &  $\mathcal{l}1$ & $2^{-6}$ & $71.60$ & $65.17$ & $0.68$ & $0.66$ \\
        \cline{2-7}
        &  $\mathcal{l}2$ & $2^{-5}$ & $69.14$ & $60.22$  & $0.64$ & $0.64$ \\
        \hline
        \multirow{2}{*}{adora$2$a} &  $\mathcal{l}1$ & $2^{-6}$ & $70.53$ & $82.72$ & $0.76$  & $0.74$ \\
        \cline{2-7}
        &  $\mathcal{l}2$ & $2^{-7}$ & $70.53$ & $83.75$ & $0.77$ & $0.74$ \\
        \hline
        \multirow{2}{*}{cnr$1$} &  $\mathcal{l}1$ & $2^{-6}$ & $90.00$ & $82.50$ & $0.86$ & $0.77$ \\
        \cline{2-7}
        &  $\mathcal{l}2$ & $2^{-6}$ & $87.27$ & $81.36$ & $0.84$ & $0.74$ \\
        \hline
        \multirow{2}{*}{cnr$2$} &  $\mathcal{l}1$ & $2^{-6}$ & $83.93$ & $82.46$ & $0.83$ & $0.72$ \\
        \cline{2-7}
        &  $\mathcal{l}2$ & $2^{-5}$ & $86.61$ & $82.91$ & $0.85$ & $0.74$ \\
        \hline
    \end{tabular}
	\label{tab:uncalibratedModelScores}
\end{table}

	Looking at performance scores presented in \Cref{tab:uncalibratedModelScores}, we can see that $\mathcal{l}1$-loss SVM outperforms $\mathcal{l}2$-loss in overall performance scores (F$1$ and AUC) in order to abl$1$ and cnr$1$ datasets, while $\mathcal{l}2$-loss SVM provides slightly better models for adora$2$a and cnr$2$ datasets. As we mentioned in Section $2$, $\mathcal{l}$1-loss SVM commonly produces better quality models. However, we have to take into an account that we relax bias $b$ in our approaches. Therefore the models are relaxed as well, which could be a cause of these unexpected results in the sense of performance score of models. 

\begin{table}[H]
	\centering
	\caption{abl$1$, adora$2$a, cnr$1$, cnr$2$ calibrated single-target models: quality of models in probabilistic sense (Brier score) and performance scores in sense of binary classification on test datasets related to models, which are converted from target probabilities to labels, using the optimal threshold (Thr.). Results are presented for the both  $\mathcal{l}1$-loss and $\mathcal{l}2$-loss SVM. }
	\begin{tabular}{| l | c | c | c | c | c | c |}
		\hline
		\multirow{3}{*}{Target} & \multirow{3}{*}{Loss} &  \multicolumn{5}{c|}{Calibrated model} \\
		\cline{3-7}
		& & \multirow{2}{*}{Brier score} & \multicolumn{4}{c|}{Binary classification} \\
		 \cline{4-7}
		& & &  Thr.  & Pre. [\%]  & Sen. [\%]  &  AUC \\
		\hline
		 \multirow{2}{*}{abl$1$} & $\mathcal{l}1$ & $0.1105$  & $0.52$ & $65.43$ &   $64.63$ & $0.64$ \\
		 \cline{2-7}
		 & $\mathcal{l}2$ & $0.0947$ & $0.54$ & $60.49$ & $60.49$  & $0.60$  \\
		\hline
		 \multirow{2}{*}{adora$2$a} & $\mathcal{l}1$ & $0.1280$ & $0.41$ & $78.95$ & $78.95$ & $0.74$ \\
		 \cline{2-7}
		 & $\mathcal{l}2$ & $0.1222$ & $0.44$ & $78.95$ & $78.95$ & $0.74$ \\
		 \hline
		  \multirow{2}{*}{cnr$1$} & $\mathcal{l}1$ & $0.0905$ & $0.63$ & $83.64$ &   $83.64$ & $0.74$ \\
		  \cline{2-7}
		  & $\mathcal{l}2$  & $0.0710$ & $0.58$  & $83.64$ & $83.64$ & $0.74$ \\
		  \hline
		   \multirow{2}{*}{cnr$2$} & $\mathcal{l}1$ & $0.0889$ & $0.53$ & $83.04$ & $83.04$ & $0.72$ \\
		   \cline{2-7}
		   & $\mathcal{l}2$ & $0.0611$ & $0.53$ & $83.93$ & $83.93$ & $0.73$ \\
		   \hline
	\end{tabular}
	\label{tab:calibratedModelScores}
\end{table}

	Analysing predictive relevance of uncalibrated models, we can see that active ligands are preferred in cases of models related to abl$1$, cnr$1$, and cnr$2$, on the other hand, inactive ligands are prefered for adora$2$a dataset. To balance predictive relevance, we perform model calibration. After this, we can see in \Cref{tab:calibratedModelScores}, the models trained using $\mathcal{l}$2-loss seem to be better calibrated by comparing Brier scores for all cases than ones related to the $\mathcal{l}$1-loss SVM. This could simple consequence of underlying model robustness. Even we use relaxed approach, the $\mathcal{l}$1-loss SVM still tries to produce a more robust model than the $\mathcal{l}$2-loss SVM, since using a linear sum of hinge loss functions instead of a sum of squared hinge loss functions leads to slightly better catching the outliers. 

\begin{table}[H]
        \centering
        \caption{abl$1$, adora$2$a, cnr$1$,  cnr$2$ biological targets: elapsed time related to training of models including HyperOpt and calibration.}
        \begin{tabular}{| c | c | c | c | c |}
            \hline
            \multirow{1}{*}{Loss} \hspace{2pt} & \multicolumn{4}{c|}{\shortstack[c]{Elapsed time [s] \\ (HyperOpt + Training + Calibration)}} \\
            \cline{2-5}
             & \hspace{2pt} abl$1$ \hspace{2pt} & \hspace{2pt} adora$2$a \hspace{2pt} & \hspace{2pt} cnr$1$ \hspace{2pt} & \hspace{2pt} cnr$2$ \\
            \hline
            $\mathcal{l}1$ & $2.15$ & $2.61$ & $1.95$ & $2.57$ \\
            \hline
            $\mathcal{l}2$ & $1.38$ & $1.86$ & $1.57$ & $1.58$ \\
            \hline
        \end{tabular}
        \label{tab:elapTimeTraining}
\end{table}

	Thus, calibrating models related to the $\mathcal{l}$2-loss SVM has significant impact than in a case of the $ \mathcal{l}$1-loss SVM; as we can see, predictive relevances of classes are well balanced. Moreover, from the \Cref{tab:elapTimeTraining}, we can observe speedups $1.56$ (abl$1$), $1.40$ (adora$2$a), $1.24$ (cnr$2$), and $1.62$ (cnr$1$) in order to using the $\mathcal{l}2$-loss SVM against the $\mathcal{l}1$-loss SVM. 
	
	However, calibrating models could cause a deterioration of overall model performance determined by means of AUC as we can see in \Cref{tab:elapTimeTraining}. Specifically, the over performance scores of models decrease by $1\%$ to $4\%$ in the cases of the abl$1$ (both loss-type models), cnr$1$ ($\mathcal{l}1$-loss model) and cnr$2$ ($\mathcal{l}2$-loss model). In order to models related to adora$2$a target, $\mathcal{l}1$-loss and $\mathcal{l}2$-loss models associated with cnr$1$ and cnr$2$, respectively, AUC scores are same. Since models were trained, calibrated and tested on different datasets, we can consider the calibrated models have the same overall performance score in the sense of AUC as their related uncalibrated models. Comparing the quality of the remaining calibrated and uncalibrated models is application-specific. In some application, they could be considered as models of same quality. For more strict quality merits, the models can be considered that differ significantly. 
		
	From achieved results, it seems that it is better to train models using the $\mathcal{l}2$-loss SVM that are not such robust as in the case of the $\mathcal{l}1$-loss SVM and, then, perform their calibration. Moreover, we can obtain a better convergence rate by employing this approach. We observe speedup up to $1.62$ in case of training model on the cnr$2$ dataset.
	
	\section{Conclusion}

	In this paper, we focused on a problem dealing with balancing predictive relevance of single-target models trained using SVMs. This calibration could be required since SVMs is sensitive to imbalanced datasets, outliers and high multicorrelation among training samples. 

	Regarding calibration improvements of models, we observe that an additional calibration works significantly better for models trained using the $\mathcal{l}2$-loss SVM with relaxed-bias from achieved results. It seems this could be a consequence of that models are not such robust as in the case of the  $\mathcal{l}1$-loss SVM. Moreover, we achieve speedup up to $1.62$ by means of the approach based on  $\mathcal{l}2$-loss SVM. On the other hand, calibrating models could cause a deterioration of overall model performance as we saw in the presented numerical experiments. Therefore, it makes sense calibrating models just for critical applications, e.g. biochemical modelling presented in this paper, where balanced predictive relevance is required.  

	Since we achieved some unexpected results in the sense of model performance scores, which are probably caused relaxing the bias term of the hyperplane $H$, we are going to focus on calibrating model trained to employ training based on full-formulation dual formulation of SVM, i.e. with equality constraint. Further, we are going to test another calibration technique, e.g. the isotonic regression.
	
	\section*{Acknowledgments}

The author acknowledge the support of The Ministry of Education, Youth and Sports from the National Programme of Sustainability (NPU II) project “IT4Innovations excellence in science - LQ1602”; the grant programme “Support for Science and Research in the Moravia–Silesia Region 2017” (RRC/10/2017), financed
from the budget of the Moravian–Silesian region; and the Grant of SGS No.
SP2020/84, VSB - Technical University of Ostrava. The author would like to thank a reviewer for the constructive feedback as well.

\bibliography{bibliography.bib}

\begin{thebibliography}{1}
\providecommand{\url}[1]{\texttt{#1}}
\providecommand{\urlprefix}{URL }

\bibitem{CorVap-ML-1995}
Cortes, C., Vapnik, V.: Support-vector networks. Machine Learning  (1995)

\bibitem{Dos-book-09}
Dost\'{a}l, Z.: Optimal Quadratic Programming Algorithms, with Applications to
  Variational Inequalities, vol.~23. SOIA, Springer, New York, US (2009)

\bibitem{Levenberg-QAM-1944}
Levenberg, K.: A method for the solution of certain non-linear problems in
  least squares. Quarterly of applied mathematics  2(2),  164--168 (1944)

\bibitem{Lin-ML-2007}
Lin, H.T., Lin, C.J., Weng, R.C.: A note on {Platt's} probabilistic outputs for
  {Support Vector Machines}. Machine Learning  68(3),  267--276 (aug 2007)

\bibitem{Pecha-SVM-AIP-2018}
Pecha, M., Hapla, V., Hor\'ak, D., \v{C}erm\'ak, M.: Notes on the preliminary
  results of a linear two-class classifier in the {PERMON} toolbox. In: AIP
  Conference Proceedings. vol. 1978 (2018)

\bibitem{Pecha-LNEE-2019}
Pecha, M., Horák, D.: Analyzing l1-loss and l2-loss support vector machines
  implemented in {PERMON} toolbox. In: Lecture Notes in Electrical Engineering,
  pp. 13--23. Springer International Publishing (Apr 2020)

\bibitem{Platt-Advances-1999}
Platt, J.: Probabilistic outputs for {Support Vector Machines} and comparisons
  to regularized likelihood methods. Advances in large margin classifiers
  10(3),  61--74 (1999)

\bibitem{Sun-JOC-2017}
Sun, J., Jeliazkova, N., Chupakhin, V., Golib-Dzib, J.F., Engkvist, O.,
  Carlsson, L., Wegner, J., Ceulemans, H., Georgiev, I., Jeliazkov, V., Kochev,
  N., Ashby, T.J., Chen, H.: {ExCAPE}-{DB}: an integrated large scale dataset
  facilitating big data analysis in chemogenomics. Journal of Cheminformatics
  9(1) (mar 2017)

\end{thebibliography}
 
\end{document}